\pdfoutput=1

\documentclass[11pt]{article}

\usepackage[final]{acl}

\usepackage{times}
\usepackage{latexsym}
\usepackage[T1]{fontenc}
\usepackage[utf8]{inputenc}
\usepackage{microtype}
\usepackage{inconsolata}
\usepackage{amsmath}
\usepackage{amssymb}
\usepackage{amsthm}
\usepackage{graphicx}
\usepackage{mathtools}
\usepackage{listings}
\usepackage{tikz}
\usepackage{siunitx}
\usepackage{tipa}
\usepackage{float}
\usepackage{bbm}
\usepackage{breqn}
\usepackage{utils}
\usepackage{tikz-dependency}
\usepackage{booktabs}
\usepackage{hyperref}
\usepackage{linguex}
\usepackage{multirow}
\usepackage{seqsplit}
\usepackage[shortlabels]{enumitem}

\usepackage[textsize=footnotesize]{todonotes}
\usepackage{xcolor}

\usepackage{caption}
\usepackage{subcaption}
\usepackage{linguex}

\newcommand{\sense}[1]{\texttt{\seqsplit{#1}}} 
\newcommand{\sensenosplit}[1]{\texttt{#1}}

\newcommand{\lrm}{Qwen2.5-DS}

\newcommand{\dataset}{{\sc Wugnectives}}

\title{\dataset: Novel Entity Inferences of Language Models from Discourse Connectives}

\author{Daniel Brubaker\ \ \ \ William Sheffield\ \ \ \ Junyi Jessy Li\ \ \ \ Kanishka Misra\\
Department of Linguistics\\
The University of Texas at Austin\\
\texttt{\{dbrubaker, sheffieldw, jessy, kmisra@utexas.edu\}}}

\begin{document}
\maketitle
\begin{abstract}

The role of world knowledge has been particularly crucial to predict the discourse connective that marks the discourse relation between two arguments, with language models (LMs) being generally successful at this task. We flip this premise in our work, and instead study the inverse problem of understanding whether discourse connectives can inform LMs about the world. To this end, we present \dataset{}, a dataset of 8,880 stimuli that evaluates LMs' inferences about \textit{novel} entities in contexts where connectives link the entities to particular attributes. 
On investigating 17 different LMs at various scales, and training regimens, we found that tuning an LM to show reasoning behavior yields noteworthy improvements on most connectives. At the same time, there was a large variation in LMs' overall performance across connective type, with \textit{all} models systematically struggling on connectives that express a concessive meaning. 
Our findings pave the way for more nuanced investigations into the functional role of language cues as captured by LMs.
We release \dataset{} at \url{https://github.com/kanishkamisra/wugnectives}
\end{abstract}

\section{Introduction}

\let\thefootnote\relax\footnotetext{\dataset{}'s name is inspired by \textit{wug}, a nonce word initially used in language acquisition research by \citet{berko1958child}.}

Discourse connectives such as \textit{but, moreover, although, because,} etc., are central to producing and comprehending natural language. As such, the task of successfully predicting a discourse connective, given two discourse arguments has been popular throughout computational linguistics research \citep{zhou-etal-2010-predicting, biran-mckeown-2013-aggregated, patterson-kehler-2013-predicting}, having made its way to the evaluation of large language models \citep{pandia-etal-2021-pragmatic, beyer-etal-2021-incoherence}. For instance, the connective that links \cref{ex:introa} to \cref{ex:introb} linearly is likely to be \textit{because} rather than \textit{although}.

\ex. \label{ex:introexample} 
\a. \label{ex:introa} I prefer Dubai to New York.
\b. \label{ex:introb} I hate snowy winters.
\vspace{-0.5em}

Prediction of the correct connective in the above example primarily requires consulting one's world knowledge---that Dubai does not have snowy winters (while New York does), and therefore \cref{ex:introb} can be a \textit{reason} for why the speaker prefers Dubai over New York. There has been a history of using such knowledge to predict discourse relations in the absence of an explicit cue \cite{marcu-echihabi-2002-unsupervised,pitler-etal-2009-automatic,lin-etal-2009-recognizing,biran-mckeown-2013-aggregated,li-nenkova-2014-reducing,rutherford-xue-2014-discovering,braud-denis-2015-comparing}.

\begin{figure}[!t]
    \centering
    \includegraphics[width=\linewidth]{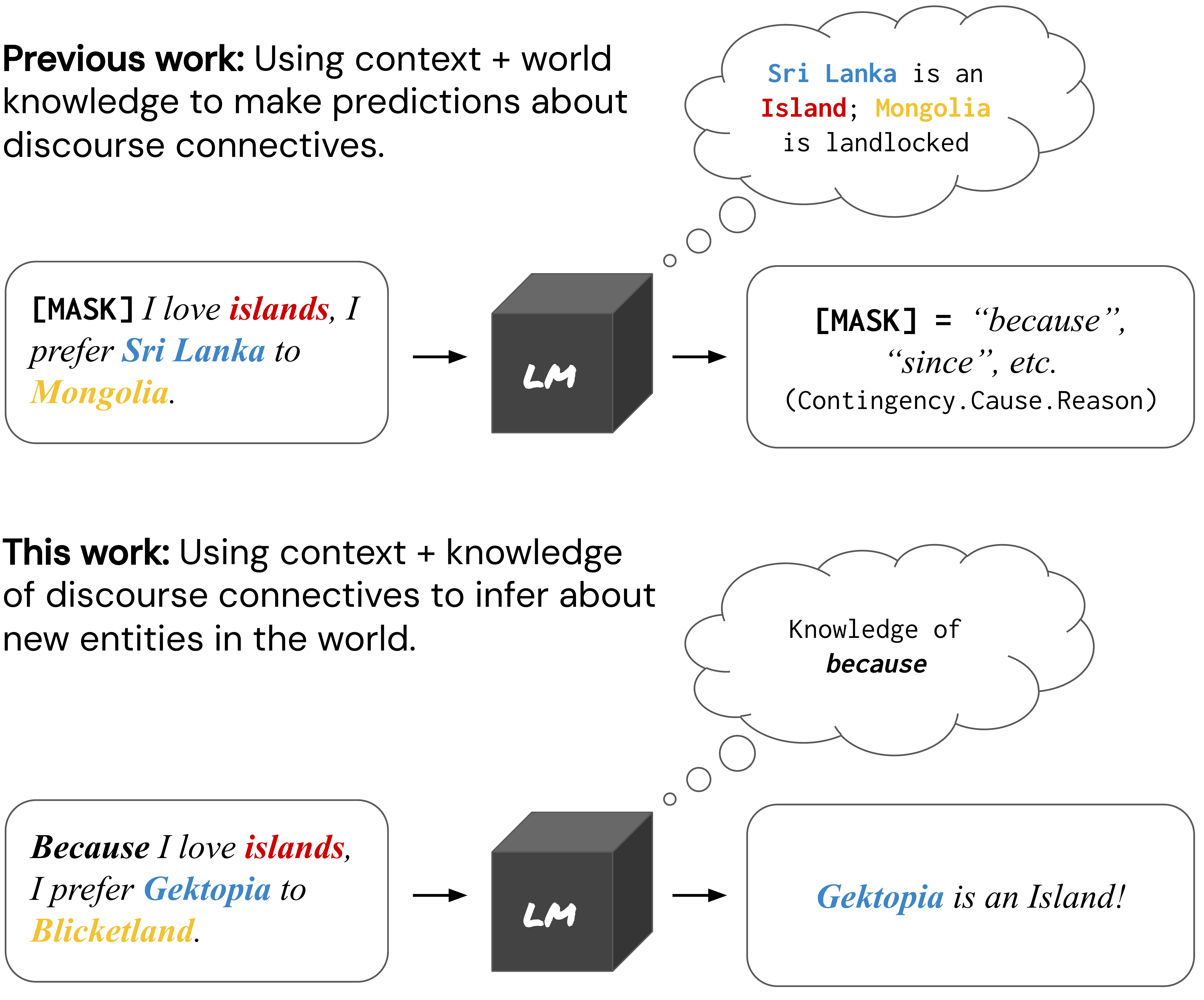}
    \vspace{-1.5em}
    \caption{Past work 
    has largely focused on the prediction of connectives given some input context, usually requiring access to world knowledge. 
    We reduce this reliance by using novel entities, and analyze whether LMs can rely on their knowledge of the connectives themselves to make inferences about the world.}
    \label{fig:fig1}
    \vspace{-0.5em}
\end{figure}

{\textbf{This paper studies the \emph{inverse} problem: what kind of inferences can one make about the entities, given the discourse connective?}}
If one were to instead consider \cref{ex:introexample2}, which replaces Dublin and New York with novel entities, X and Y, then depending on the connective used, one could draw different inferences about what X and Y could be.

\ex. \label{ex:introexample2} 
\a. \label{ex:intro2a} I prefer X to Y.
\b. \label{ex:intro2b} I hate snowy winters.

For instance, if we were to use \textit{because}, then one could infer that X does not have snowy winters. On the other hand, if we use \textit{however}, then this inference is reversed---that despite hating snowy winters the speaker prefers X (over Y). 
In this manner, discourse connectives can serve as \textit{cues} to meaning/world knowledge \citep{elman2004alternative}.

{In many ways, modern LLMs have been regarded as models that have finally ``grasped'' language \citep{piantadosi2023modern, futrell2025linguistics}. Their use of discourse connectives in modern AI-generated writing has similarly progressed far from older LMs \cite{ko-li-2020-assessing}. But there is a difference between being able to \emph{use} connectives correctly given known properties of concepts involved, \textit{v.s.}~fully \emph{understand} their functional meaning and making correct \emph{inferences} \citep{mahowald2024dissociating}.}
We ask: \textbf{to what extent do LMs make abstract inferences about novel entities as licensed by connectives?} Answering this question allows us to make fine-grained contributions to the broader goal of characterizing how well aspects of meaning---in this case, attributes and relations of novel entities---can be learned from language exposure \citep{gelman2004learning, lupyan2019words}.

To answer this question, we propose \dataset{}, a benchmark consisting of 740 utterances, each of which links novel entities to their attributes via the usage of specific discourse connectives. These utterances are embedded across 12 different prompt variations, amounting to 8,880 unique stimuli. 
Thus, while existing
work focusing on discourse connectives effectively tests how world knowledge enables the prediction (or comprehension) of discourse connectives, 
this work flips this premise,
and instead investigates how 
specific connectives can inform the model about entities mentioned in the arguments that the connectives operates over. By removing the aspect of world knowledge that is likely entrenched in  
a language model (LM)'s
parameters, 
our investigation sheds light on whether LMs learn the \textit{abstract} functional meaning of these connectives, in a manner that is independent of the content of the arguments they connect.

\dataset{} consists of stimuli for 41 unique connective-forms, spanning 7 different senses, across a total of 3 different stimuli types, each focusing on different kinds of knowledge about novel entities---e.g., instantiation/category membership, temporal relations, and general attributes of entities such as ``\textit{being an island nation}''. Using \dataset{}, we evaluate 17 different open-source LMs at various parameter sizes, and training types (base, instruction tuning, and reasoning-based tuning).

We find LMs to show a great deal of variation in their abilities to infer about entities from connective usage. LMs generally performed above chance on cases where connectives expressed temporal meaning between novel events, or provided causal evidence for an entity's attributes (or lack thereof), and in some limited cases, when they expressed instantiation relations between entities. However, they consistently obtained chance-level performance on connectives that expressed \textit{concession} between arguments---i.e., when a causal expectation raised on the basis of one argument is denied by the other. The systematicity of our results on such cases suggested that these classes of connectives (\textit{although, even though, despite that}, etc.) pose a fundamental difficulty for LMs to reason about the semantic features of novel entities in context. Our results could not be explained by frequency of these connectives in internet corpora, suggesting a more nuanced, intractable reason at play. Finally, while we found no clear effect of scale or instruction-tuning, we did find reasoning-based tuning of LMs \citep[\'a la][]{guo2025deepseek} to be beneficial, achieving the best performance across all connective senses (though still struggling on concession). Lastly, we ran a post-hoc experiment where models were evaluated on stimuli with real-world entities---ones grounded in world knowledge---in place of novel entities. As expected, we found significantly increased performance, though this increase was weakest for concessive connectives, reinforcing the conclusion that their inferential knowledge is especially difficult for LMs to capture.
In sum, our findings pave the way for more nuanced investigations into the functional meanings of language cues, complementing traditional usage-based analyses.

\section{Background}

\paragraph{Modeling Connectives}

Discourse connectives, e.g., ``because'', ``however'', etc., are a class of words that mark the discourse (coherence) relations between two arguments, often unambiguously \citep{pitler-nenkova-2009-using}.~The role of world knowledge contained in the arguments of a relation has been of particular significance in prior work on recognizing implicit discourse relations. While this was initially signified using cartesian products of words in arguments \citep{marcu-echihabi-2002-unsupervised}, there was widespread operationalization of this premise in several works since the introduction of the Penn Discourse Treebank \citep[PDTB;][]{prasad2017penn}---e.g., \citet{pitler-etal-2009-automatic,lin-etal-2009-recognizing,zhou-etal-2010-predicting,biran-mckeown-2013-aggregated, patterson-kehler-2013-predicting, li-nenkova-2014-reducing,rutherford-xue-2014-discovering,braud-denis-2015-comparing}. \citet{braud-denis-2016-learning} first showed evidence of improved discourse relation classification when word representations were informed by connectives, and \citet{miao-etal-2024-discursive} showed that LMs metalinguistic discursive reasoning, as measured by \textsc{DiSQ} scores, is improved when explicit discourse connectives are present.
Previous work evaluating language models' understanding of discourse connectives has largely focused on their ability to categorize, or predict connectives in context \citep{nie-etal-2019-dissent, kim-etal-2020-implicit, koto-etal-2021-discourse}. 
A number of works, such as \citet{pandia-etal-2021-pragmatic}, CoherenceGym \citep{beyer-etal-2021-incoherence}, and \citet{cong-etal-2023-investigating} have evaluated LMs' sensitivity to infelicitous usage of connectives. 
A common theme among these works is that world knowledge about the arguments being linked together is a necessary prerequisite to models' success. Our work abstracts away from this assumption by testing how connectives \textit{cue} the meanings of the arguments that they link.

\paragraph{Learning about the world from Language (models)}
LMs offer an interesting avenue to investigate questions about how language exposure can give rise to semantic knowledge---a subject that has always received great theoretical interest \citep{landau1985language, waxman1995words, elman2004alternative, gelman2004learning, lupyan2019words}. Grounded in this motivation, a number of previous works have aimed to characterize the kinds of world knowledge that arises in LMs \citep[][i.a.]{abdou-etal-2021-language, misra-etal-2023-comps, ivanova2024elements}. While these works serve as catalogs of what kinds of world knowledge can be acquired from language exposure, the status of the cues that can give rise to them is less clear. 
Nonetheless, there have been a range of proposals about templates with specific discourse connectives serving as cues to broader meaning, primarily in terms of paradigmatic relations \citep{hearst1992automatic, jones2003antonymy, murphy2009discourse, roth-schulte-im-walde-2014-combining, zhang-et-al-2020-ASER}. For instance, templates like ``\textit{Xs} such as \textit{Ys}'' suggests a hypernymy relation between \textit{X} and \textit{Y} \citep{hearst1992automatic}. Similarly, connectives like \textit{but} can be indicative of antonym relations \citep{jones2003antonymy, murphy2009discourse}. Our work sheds light on the extent to which cues similar to these could give rise to semantic knowledge about novel entities in LMs, under limited exposures.

\label{sec:stimuli_design}
\section{Designing \dataset{}}

In this section we describe the design decisions for \dataset{}---in terms of how we operationalize ``novel information'', type of inference, and choice of connectives, finally culminating in our description of the stimuli.

\paragraph{Nonce words} To operationalize `novel information' we use nonce words as novel entities in utterances expressing a proposition from which the LM has to infer their attributes (e.g., \textit{is a leafy vegetable}) or their relations to another novel entity (e.g., \textit{Wugs are Daxes}). The usage of nonce words for reasoning has now become commonplace in computational linguistics research \citep{misra-etal-2023-comps, eisenschlos-etal-2023-winodict, rodriguez-etal-2025-characterizing}, and has been a longstanding tradition in cognitive psychology to mimic a scenario where the learner has little, if any, knowledge of the entity/property in question \citep{osherson1990category, gelman2010effects}.  Following \citet{misra-etal-2023-comps}, we use pairs of nonce words in our stimuli, primarily due to two reasons. First, a number of our chosen connectives (see below) describe relations between two entities, and therefore, to maintain uniformity we use two novel entities throughout. Second, using two nonce words creates a notion of choice, and prevents the scenario where the LM simply uses co-occurrence information to make judgments about the only novel entity in question. Importantly, we counterbalance our nonce words throughout, to ensure that a systematic bias towards any one surface form leads to poor performance. Our novel entities range from events (for temporal connectives), to simple bare plurals (for instantiation, comparison, and contingency connectives) to locations (for comparison and contingency connectives). See \Cref{tab:wugs} for the full list of nonce words.

\begin{table}[!t]
\centering
\resizebox{\columnwidth}{!}{%
\begin{tabular}{@{}ll@{}}
\toprule
\textbf{Entity type} & \textbf{Surface forms} \\ \midrule
Bare plurals & \textit{\begin{tabular}[c]{@{}l@{}}Wugs, Daxes, Feps, Geks, Blickets\end{tabular}} \\ 
Events & \textit{\begin{tabular}[c]{@{}l@{}}Wugfest, Daxday, Fepfestival, Gextravaganza, Blicketbash\end{tabular}} \\
Locations & \textit{\begin{tabular}[c]{@{}l@{}}Wugsville, Daxburgh, Fepopolis, Gektopia, Blicketland\end{tabular}} \\ \bottomrule
\end{tabular}%
}
\caption{Lists of nonce words used in our stimuli.}
\label{tab:wugs}
\vspace{-1em}
\end{table}

\paragraph{Inference type}

Our stimuli are designed so that the most salient inferences are \emph{entailments}, which means that incorrect responses cannot come from reasonable cancellations.
This allows us to strictly view answers that do not match the proscribed inferences as incorrect. We make the natural assumption that the speaker is using the connective and nonce words cooperatively, and knows specific meaning of the nonce words, though the listener does not. While additional context could potentially weaken these inferences, the stimuli are presented in naturalistic prompt templates that strongly bias the inferences towards entailed readings.\footnote{{In particular, one could imagine contexts in our which our stimuli with \sense{Concession} connectives have indirect-concessive readings, and these entailments do not arise. In the absence of these contexts, the direct reading (with licensed entailment) is most salient.}} We use these entailments to operationalize the notion of ground truth for all stimuli.

\begin{table*}[]
\centering
\resizebox{0.9\textwidth}{!}{%
\begin{tabular}{@{}lllll@{}}
\toprule
\textbf{Stimuli Type} & \textbf{Target Properties} & \textbf{Sense} & \textbf{Count} & \textbf{Connectives} \\ \midrule
Instantiation & \textit{is a} & \begin{tabular}[c]{@{}l@{}}\texttt{Expansion.Instantiation.}\\ \texttt{Arg2-as-instance}\end{tabular} & 100 & \textit{\begin{tabular}[c]{@{}l@{}}for example, for instance, \\ in particular, specifically, such as\end{tabular}} \\ \midrule
\multirow{4}{*}{Preference} & \multirow{4}{*}{\textit{\begin{tabular}[c]{@{}l@{}}are leafy vegetables,\\ are mammals, are fruits,\\ are string instruments,\\ are insects, is an island,\\ is a college town, \\ is a coastal city,\\ has mountains nearby,\\ has an equatorial climate\end{tabular}}} & \texttt{Contingency.Cause.Reason} & 160 & \textit{as, because, for, since} \\ \cmidrule(l){3-5} 
 &  & \texttt{Contingency.Cause.Result} & 240 & \textit{\begin{tabular}[c]{@{}l@{}}as a result, for example, for instance, \\ so, therefore, thus\end{tabular}} \\ \cmidrule(l){3-5} 
 &  & \begin{tabular}[c]{@{}l@{}}\texttt{Comparison.Concession.}\\ \texttt{Arg1-as-denier}\end{tabular} & 98 & \textit{\begin{tabular}[c]{@{}l@{}}although, as much as, \\ even though, though\end{tabular}} \\ \cmidrule(l){3-5} 
 &  & \begin{tabular}[c]{@{}l@{}}\texttt{Comparison.Concession.}\\ \texttt{Arg2-as-denier}\end{tabular} & 302 & \textit{\begin{tabular}[c]{@{}l@{}}although, but, even though, however, yet,\\ nevertheless, though, while, despite that\end{tabular}} \\ \midrule
\multirow{2}{*}{Temporal} & \multirow{2}{*}{\textit{started earlier}} & \begin{tabular}[c]{@{}l@{}}\texttt{Temporal.Asynchronous.}\\ \texttt{Precedence}\end{tabular} & 110 & \textit{\begin{tabular}[c]{@{}l@{}}afterwards, as a result, before, consequently, even before, \\ finally, hence, later, next, so, subsequently, then\end{tabular}} \\ \cmidrule(l){3-5} 
 &  & \begin{tabular}[c]{@{}l@{}}\texttt{Temporal.Asynchronous.}\\ \texttt{Succession}\end{tabular} & 130 & \textit{\begin{tabular}[c]{@{}l@{}}therefore, after, as soon as, because, earlier, even after, \\ eventually, even though, once, previously, since, thereafter\end{tabular}} \\ \bottomrule
\end{tabular}%
}
\caption{Breakdown of our stimuli, showing stimuli type, the properties they target, number of unique stimuli, possible senses included in stimuli type, and the list of connectives used. \textbf{Note:} connectives that repeat are either poly-functional (e.g., while \textit{because} is primarily a \texttt{Contingency} connective, it can also have \texttt{Temporal} senses), or have subtly different level 3 senses---e.g., whether \textit{although} appears in the clause with the preference (``I prefer wugs to daxes'') or a property (``I hate leafy vegetables'') changes its \texttt{Comparison.Concession} sense). Connective senses were determined by hand-annotating each stimuli in context.}
\label{tab:stats}
\end{table*}

\paragraph{Connectives and relation senses}
We select a subset of relations from PDTB \citep{prasad2017penn} to build our stimuli from the following level 2 senses: \sense{Expansion.Instantiation}, \sense{Contingency.Cause}, \sense{Comparison.Concession}, and \sense{Temporal.Asynchronous}. A full list of the third-level senses and connectives can be found in \Cref{tab:stats}.

\begin{table*}[ht]
\centering
\resizebox{0.8\textwidth}{!}{%
\begin{tabular}{@{}lll@{}}
\toprule
\textbf{Stimuli Type} & \textbf{Stimuli Example} & \textbf{Answer} \\ \midrule
Instantiation & \textit{\begin{tabular}[c]{@{}l@{}}Daniel said, "I find feps \underline{such as} blickets, to be awful." Can we conclude\\  from this that blickets are feps? Answer either with Yes or No.\end{tabular}} & Yes \\
Preference & \textit{\begin{tabular}[c]{@{}l@{}}Lucy said, "\underline{Even though} I hate leafy vegetables, I prefer feps to geks." From \\ this, is it true that feps are leafy vegetables? Answer either with Yes or No.\end{tabular}} & Yes \\
Temporal & \textit{\begin{tabular}[c]{@{}l@{}}Erica said, ``blicketbash occurred \underline{before} gextravaganza.'' From this, which event \\ started first? Answer either with blicketbash or gextravaganza and nothing else.\end{tabular}} & blicketbash \\ \bottomrule
\end{tabular}%
}
\caption{Examples of stimuli per stimuli type, along with their answers.}
\label{tab:stimex}
\vspace{-0.5em}
\end{table*}

\paragraph{Stimuli Design} Each individual stimulus consists of two parts: a premise and inference. The premise is a sentences which uses a discourse connective to specify the relation between two nonce words. The inference is either an explicit formulation of that entailed relation, its logical opposite, i.e, a contradiction. Models are prompted with each inference framed as a question, and must respond appropriately based on whether the inference is entailed or contradictory. We categorize these stimuli into three families based on the types of inferences they license: Temporal, Instantiation, and Preference.

\textbf{\textit{Instantiation:}} The connectives in these stimuli describe the \textbf{is-a} relationship between two discourse entities. Stimuli of this type follow the form of sentence \cref{ex:insta}, and systematically entail sentences in the form of \cref{ex:instb}. 

\ex. \label{ex:instantiationexample}
 \label{ex:insta} I like wugs, \textit{for example}, daxes are nice.

\ex. \label{ex:instantiationexample2}
\label{ex:instb} Daxes are wugs.

We used five \sense{Expansion.Instantiation.Arg2-as-instance} connectives in these stimuli.

\textbf{\textit{Preference:}} Here, connectives describe the relationship between a speaker's preferred entity and a liked or disliked property. These inferences can either be causal or concessive, in which case the licensed inferences respectively arise because of or in spite of expectations raised by the premise. For the purposes of stimuli design, connectives in this category are most saliently organized by their level-2 senses, \sense{Contingency.Cause} (like \textit{because}) and \sense{Comparison.Concession} (like \textit{although}).\footnote{However, there is variety among third-level senses. See \Cref{tab:stats} for more details.} For example, both \cref{ex:prefa} and \cref{ex:prefb} entail \cref{ex:prefc}.

\ex. \label{ex:preferenceexample}
\a. \label{ex:prefa} \textit{Because} I love leafy vegetables, I prefer wugs to daxes.
\b. \label{ex:prefb} \textit{Although} I hate leafy vegetables, I prefer wugs to daxes.

\ex. \label{ex:prefc} Wugs are leafy vegetables.

 Additionally, the use of ‘love’ and ‘hate’ can be swapped to change the polarity of the entailment. That is, the first nonce\footnote{Inferences about the second entity's relation to the property, while occasionally salient, are indeterminate in their entailment status and were subject of much debate amongst the authors. We do not make any claims about whether or not these inferences are implicatures or entailments, and leave them out of our dataset.
}
 will not have the property, as in \cref{ex:prefd} and \cref{ex:prefe}, both of which entail \cref{ex:prefg}.\footnote{
    Notably, this being an entailment rather than an implicature depends on the property being most saliently being binary. Consider the following example with a gradable property: ``Although I love tall buildings, I prefer Austin to New York. Austin does have tall buildings, just not as many as New York.'' The second sentence cancels the inference, indicating it is an implicature rather than an entailment.
 }

\ex. \label{ex:preferenceexample2}
\a. \label{ex:prefd} \textit{Because} I hate leafy vegetables, I prefer wugs to daxes.
\b. \label{ex:prefe} \textit{Although} I love leafy vegetables, I prefer wugs to daxes. 

\ex. \label{ex:prefg} Wugs are \textbf{not} leafy vegetables.

The questions regarding inferences where the nonce does not have the property (such as \cref{ex:prefg}) are identical to those where the property applies, but the correct answer changes from ``Yes'' to ``No.''
This is the general form of preference stimuli, though a few other variations of these stimuli are present as well. We use both fronted and non-fronted connectives where naturally possible. Additionally, the connective can occur in the clause with the preference rather than the property while still licensing the same inference (e.g. ``\textit{Although} I prefer wugs to daxes, I hate leafy vegetables.'', which still entails \cref{ex:prefg}). 

Overall, we use 20 unique connectives for these stimuli, with 10 each in \sense{Contingency.Cause} and \sensenosplit{Comparison.Concession} senses (\Cref{tab:stats}). We use a total of 10 unique entity properties---5 mapped to bare plurals, and 5 mapped to locations. 

\textbf{\textit{Temporal:}} Here, connectives describe the temporal order of events. We include connectives from both \sense{Temporal.Asynchronous.Precedence} \cref{ex:tempa} and \sense{Temporal.Asynchronous.Succession} \cref{ex:tempb}. 

\ex. \label{ex:temporalexample} 
\a. \label{ex:tempa} Wugfest happened \textit{even before} Daxday took place.
\b. \label{ex:tempb} Daxday happened \textit{once} Wugfest took place.

Both sentences \cref{ex:tempa} and \cref{ex:tempb} entail \cref{ex:tempc}.

\ex. \label{ex:temporalexample2} \label{ex:tempc} Wugfest started before Daxday.

More explicitly, \sense{Precedence} connectives license inferences where the entity in Arg1 starts before that in Arg2, and \sense{Succession} connectives license inferences where the entity in Arg2 starts before Arg1. This does not directly map to the first entity linearly, as fronting the connective changes the order of the underlying arguments while preserving the licensed inference, as in \cref{ex:temp-front}.

\ex. \label{ex:temporalexample3} \label{ex:temp-front} \textit{Once} Wugfest took place, Daxday happened.

Additionally, we randomly vary the verbs used to say that each event came to pass between \textit{happened}, \textit{took place}, and \textit{occurred}. These verbs are neutral with respect to a start time, and the increased variety ensures both naturalistic stimuli and helps isolate connective-based pragmatic reasoning from over-reliance on the surface form. Unlike Instantiation and Preference stimuli, the inferences in this family are prompted in open-ended questions (e.g., ``Which event started first?''). Accordingly, this category does not contain contradictory inferences, and the responses we measure from models are the names of the two given events. We use a total of 24 temporal connectives, 12 each in \sense{Temporal.Asynchronous.Precedence} and \sense{Temporal.Asynchronous.Succession}.

 \paragraph{Prompt Templates}
We embed our premise and inference pairs in 12 different prompt templates per stimuli type. Our templates are formatted in terms of a narrative dialogue where a named speaker says the premise, which is then followed by a question that targets the inference, and an instruction to guide the model to possible answers (Yes/No for Instantiation and Preference, or the name of one of the two events for Temporal). 
Variation in our prompts are paraphrases of the questions---e.g., replacing ``From this, which event started first?'' with ``From what Erica said, which of the two events began first?'' (see \cref{tab:stimex}). A detailed list of prompt templates is shown in \cref{tab:prompt-variation-pref-inst} and \cref{tab:prompt-variation-temporal} in the Appendix. After applying our prompt templates, we end up with a total of 8,880 stimuli.

\begin{figure*}[!t]
    \centering
    \includegraphics[width=\textwidth]{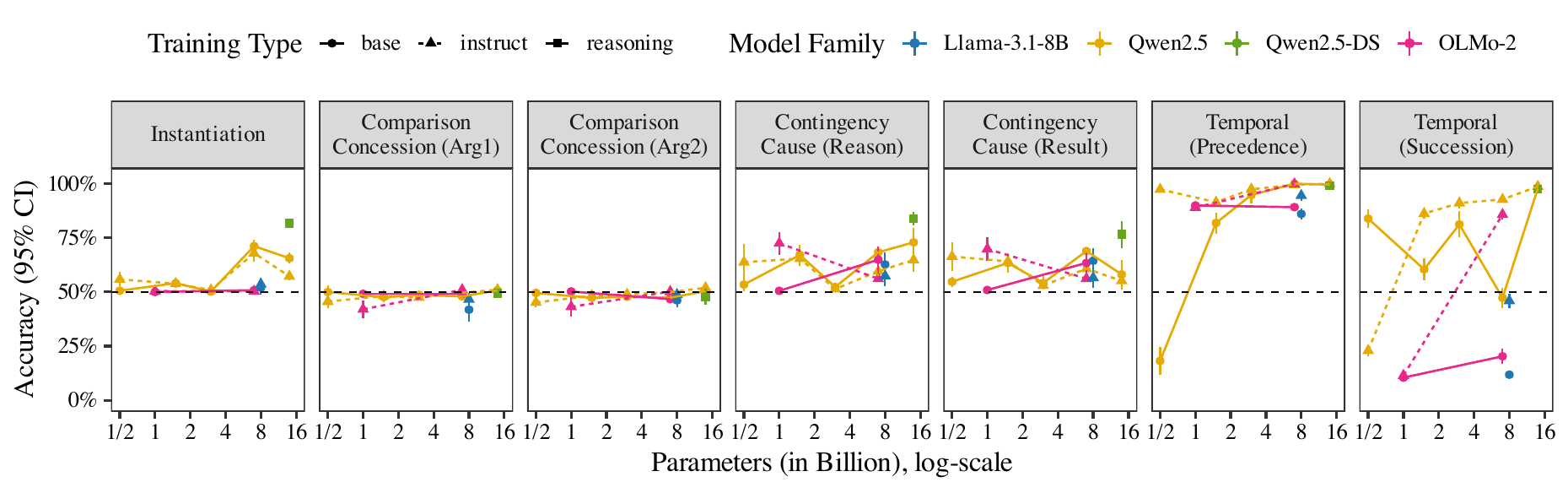}
    \vspace{-1.8em}
    \caption{Accuracy of LMs across connective senses. 
    The black dashed line indicates chance performance (50\%). Error bars indicate 95\% confidence intervals measured across connectives and prompt variation.
    }
    \label{fig:overall}
    \vspace{-0.5em}
\end{figure*}

\section{Experimental Setup}
\label{sec:expsetup}

\paragraph{Models} We evaluate on three primary model families: Qwen-2.5 \citep{qwen2.5}, OLMo-2 \citep{olmo2}, and Llama 3.1 \citep{llama3}, across multiple different scales (when possible) in terms of parameter counts. For Qwen-2.5, we evaluate at 5 different scales ranging from 500M---14B parameters, for OLMo-2 we evaluate its 1B and 7B variants, and for Llama 3.1, we evaluate on its 8B variant. For each model we include its instruction tuned as well as non-instruction tuned (which we refer to as ``base'') versions. Additionally we also evaluate on a distilled version of the DeepSeek R1 model \citep{guo2025deepseek}, where Qwen-2.5-14B was fine-tuned on reasoning traces generated from the larger DeepSeek R1 model. We refer to this model as \lrm{}. In total, we test on 17 models: 10 Qwen-2.5 LMs, 4 OLMo-2 LMs, 2 Llama LMs, and 1 \lrm{} LM. 
\Cref{tab:model-metadata} (appendix) shows metadata of each model.

\paragraph{Answer Extraction}
For base and instruction tuned models, we extracted the answer following recent work \citep{rodriguez-etal-2025-characterizing}: in the case of preference and instantiation stimuli, since we had a Yes/No question in our stimuli, we extracted the probabilities of a variety of surface forms for Yes and No (i.e., case variation and space prefixing), and normalized them to get the relative probabilities of `Yes' and `No'. We then chose the form with the greatest probability as the model's response. We performed the same process for our temporal stimuli, but retrieved the probabilities of the pair of `Event' nonce words (see \Cref{tab:wugs}) instead. For \lrm{},~we prompt the LM to generate its responses and include its answer in the standard \texttt{\textbackslash{boxed\{\}}} format. There were cases where the LM did not do this, for which, we used the pipeline described in \Cref{sec:appendix-response-extraction} to extract predictions.

\paragraph{Measurement}
We primarily report accuracy as our main performance metric, calculated as the proportion of time the correct response was produced by the model using our extraction strategy described above. Across all levels of our analyses (overall, per-sense, per connective, etc.), we report 95\% confidence intervals across different prompt templates to jointly characterize the effect of prompt variation. Since in all cases we evaluate models on their choice between two possible answers (Yes/No for Preference and Instantiation stimuli, and between two entities for Temporal stimuli), chance accuracy is 50\%.\footnote{For temporal stimuli, we counterbalance the nonce words across all possible pairs such that a bias towards one will result in chance performance.}

\section{Results and Analysis}

We analyze our results along two main threads. We start by focusing on model performance by connective sense, diving deeper into salient patterns of model behavior on specific connectives or a class of connectives. We then turn to analyses that focus on external artifacts such as model scale, factors involved in the models' training regimen such as instruction tuning, or training models to perform reasoning (in the sense of DeepSeek-R1), and an analysis of model performance vs. connective frequency. Our main results across all these variables, broken down by sense, is shown in \Cref{fig:overall}.

\subsection{By Discourse Sense}

Only a few LMs captured entailments licensed by connectives across senses.~While LMs generally had above-chance accuracies on \sense{Temporal}, \sense{Contingency}, and in some cases, \sensenosplit{Instantiation} connectives, they consistently struggled on \sensenosplit{Concession}, suggesting a systematic lack of abstract knowledge of this particular sense.
Below we discuss more detailed results per sense:

\begin{figure}[!t]
    \centering
    \includegraphics[width = \columnwidth]
    {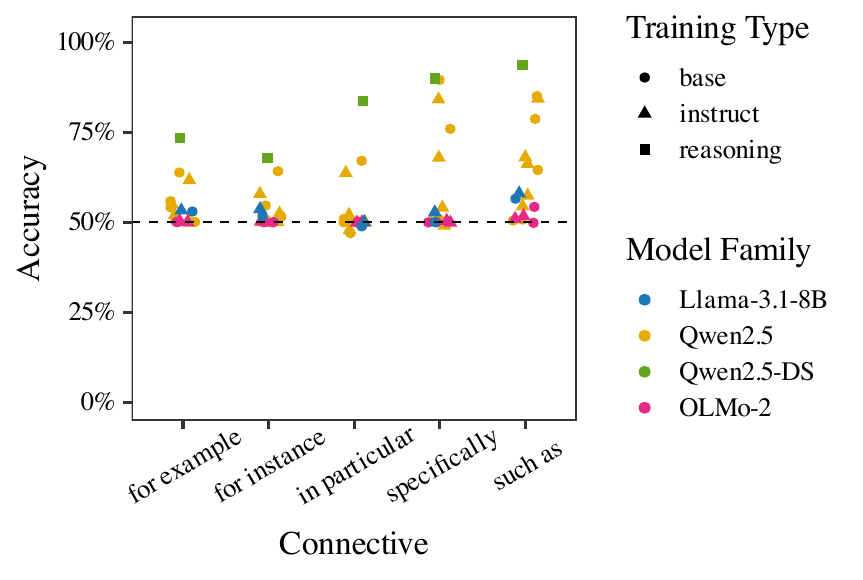}
    \vspace{-2em}
    \caption{Mean accuracy (across prompts) of models by connective for the sense \texttt{Expansion.Instantiation}.}
    \label{fig:inst}
    \vspace{-0.5em}
\end{figure}

\paragraph{Instantiation}

Apart from a select few cases (i.e., Qwen-2.5 models at or above 7B parameters), most LMs were at chance performance on Instantiation connectives, with the \lrm{} model performing the best (at 82\% accuracy). \Cref{fig:inst} shows results of models broken down by connective. We observe a notable amount of variation across instantiation connectives, with models particularly struggling on \textit{for instance} (avg. accuracy of 53\%) and \textit{for example} (avg. accuracy of 63\%).

\paragraph{Concession}

\begin{figure}[t]
    \centering
    \includegraphics[width = 0.8\columnwidth]
    {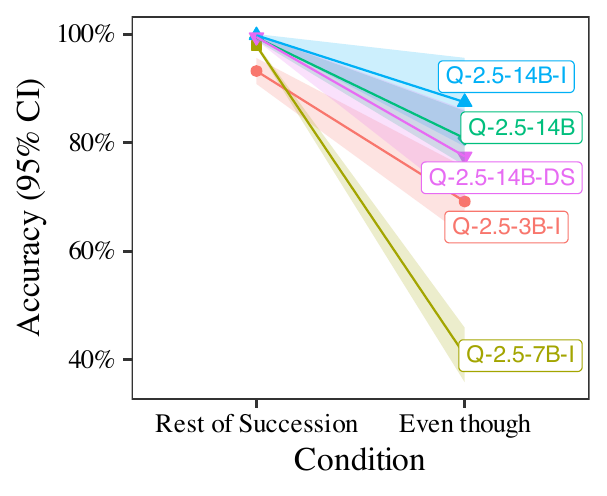}
    \vspace{-1em}
    \caption{Accuracy of top five models on succession stimuli without \textit{even though} compared with their performance on \textit{even though}. Model names are abbreviated to save space. Q: ``Qwen'', I: ``Instruct''.}
    \label{fig:even-though}
    \vspace{-0.5em}
\end{figure}

All LMs in our experiments obtained chance-level performance on Concession connectives (\texttt{Comparison.Concession}). That is, they seem to fundamentally struggle to infer entailments in cases where the function of the connective is to cancel or deny a causal relation expressed in one of the arguments. We observe this for both types of concession senses (\sense{Arg1-as-denier} as well as \sense{Arg2-as-denier}), suggesting a generally robust trend. Among various individual connectives, LMs seemed to especially struggle at \textit{although} and \textit{even though} when they are fronted (e.g., ``\textit{Although} I hate leafy vegetables, I prefer wugs to daxes.'', where \textit{although} is used in an \texttt{Arg1-as-denier} sense), oftentimes even obtaining below-chance performance. See \Cref{fig:comparison-breakdown} in the appendix for a full breakdown. In fact, models appear to struggle with concession connectives so much so that this seems to transfer over to when they are used in a different sense. Evidence for this is shown in \Cref{fig:even-though}, where we see that the performance of the top performing LMs on succession connectives is compellingly greater than that on the succession sense of ``\textit{even though}'', a connective that also has a concession sense.\footnote{As an example of \textit{even though} in a succession sense, consider: ``Wugfest happened even though daxday took place.''} This reinforces the finding of LMs' notable weakness on concession connectives. This to some extent also relates to the general processing difficulty of concession connectives (relative to causal and continuous connectives) from ERP studies on humans \citep{brehm2005connective, kohne2013time, kohne2021online}.

\paragraph{Contingency} Models are generally better on contingency connectives than on the previous two, with 13/17 models being at least 5 percentage points above chance on both types of contingency connectives. At the same time, they are considerably far from perfect, with only the \lrm{} model showing accuracies above 75\% for both senses.

\begin{table}[]
\centering
\begin{tabular}{@{}lcc@{}}
\toprule
\textbf{} & \multicolumn{1}{l}{\sense{\textbf{Precedence}}} & \multicolumn{1}{l}{\sense{\textbf{Succession}}} \\ \midrule
\textbf{Choose-first} & 92.5\% & 56.4\% \\
\textbf{Choose-recent} & 7.5\% & 43.6\% \\ \bottomrule
\end{tabular}
\caption{Performance of positional-based heuristics. Choose-first linearly selects the first entity, and choose-recent selects the latest entity.}
\label{tab:temp-position}
\vspace{-0.5em}
\end{table}

\paragraph{Temporal} Models showed largely inconsistent behavior in their performance on Temporal connectives, especially when observing the changes between their performance in the \texttt{Precedence} vs. \texttt{Succession} senses.~A few exceptions to this trend were the \lrm{} model as well as larger variants of the instruction tuned Qwen-2.5 and OLMo models. In all other cases, models showed the greatest variability in their performance relative to that on other closely related senses discussed before. 

One explanation for this variability can come from reasoning about shallow heuristics a system might potentially rely on in ``solving'' the inference problem in these stimuli. Here we shed light on two such heuristics, both heavily dependent on linear position of entities in the LMs' input: 1) \textbf{Choose-First:} select the first entity in the premise as the answer, and 2) \textbf{Choose-Recent:} select the most recent entity in the premise. These heuristics do not have the same impact on the two Temporal senses, as seen from their accuracies in \Cref{tab:temp-position}, suggesting different levels of difficulty for the two types of stimuli. For \texttt{Precedence} stimuli, applying the \textbf{Choose-first} heuristic can result in perfect performance for almost all connectives. This is because, except for 2 connectives (\textit{before}, and \textit{even before}), none of the other 10 \texttt{Precedence} connectives can be fronted---i.e., they always have to follow the same linear order of \texttt{\{event1\} <connective> \{event2\}}, and so simply extracting \texttt{\{event1\}} can result in the appearance of sophisticated reasoning. Since most \texttt{Succession} connectives can easily be fronted, neither heuristic seems to have a non-trivial role to play. Overall, it is difficult to determine if these heuristics are borne out in the LMs we analyzed, unless we perform a causal analysis of their mechanisms \citep{geiger2021causal}, which we leave for future work.

\subsection{By external artifacts}

In general, frequency alone does not explain the variance in the model's performance, and there were no clear effects of scale or instruction tuning. Though preliminary, we did find reasoning-based tuning (i.e., the manner in which \lrm{} was tuned) to consistently show stronger performance, with the exception on \texttt{Concession} connectives as discussed in the previous subsection. Below we discuss these in greater detail. We make use of linear-mixed effects regression (LMER) for our analysis of Scale, Instruction-Tuning, and Reasoning-based tuning, and report overall results here, while leaving particular details of the analysis in \Cref{sec:lmer}.

\paragraph{Frequency of Connective}

While discourse connectives are generally quite frequent in corpora, to what extent does their frequency relate to LMs' behavior in capturing their licensed inferences? To test this, we extract frequencies of our 41 unique connectives from Dolma corpus \citep{soldaini-etal-2024-dolma}, and measure their correlation with LMs' performance per connective. We do not find a significant correlation between frequency and model performance (see \Cref{tab:freq-r-squared} in the Appendix).\footnote{As a caveat, it is intractable to track what sense of a connective was being used in a corpus as large as Dolma, and therefore these frequencies can be seen as the estimates for the upper-bound of those used in their actual, precise senses.}

\paragraph{Scale} 
Overall, we do not find any notable, global effect of scale in our results. In many cases, larger models of the same family (OLMo and Llama) were no different than their smaller counterparts, while in other cases there were only selective instances of a clear effect of scale---e.g., Qwen-2.5 Instruct models on \texttt{Precedence} and Qwen-2.5 base models on \texttt{Succession}. Results from LMER analysis with an interaction term for number of parameters and sense as fixed effects, with random effects for model and prompt templates corroborated our findings ($\beta_{\text{params}} = 0.008, p = .10$).

\begin{figure*}[!t]
    \centering
    \includegraphics[width=0.8\textwidth]{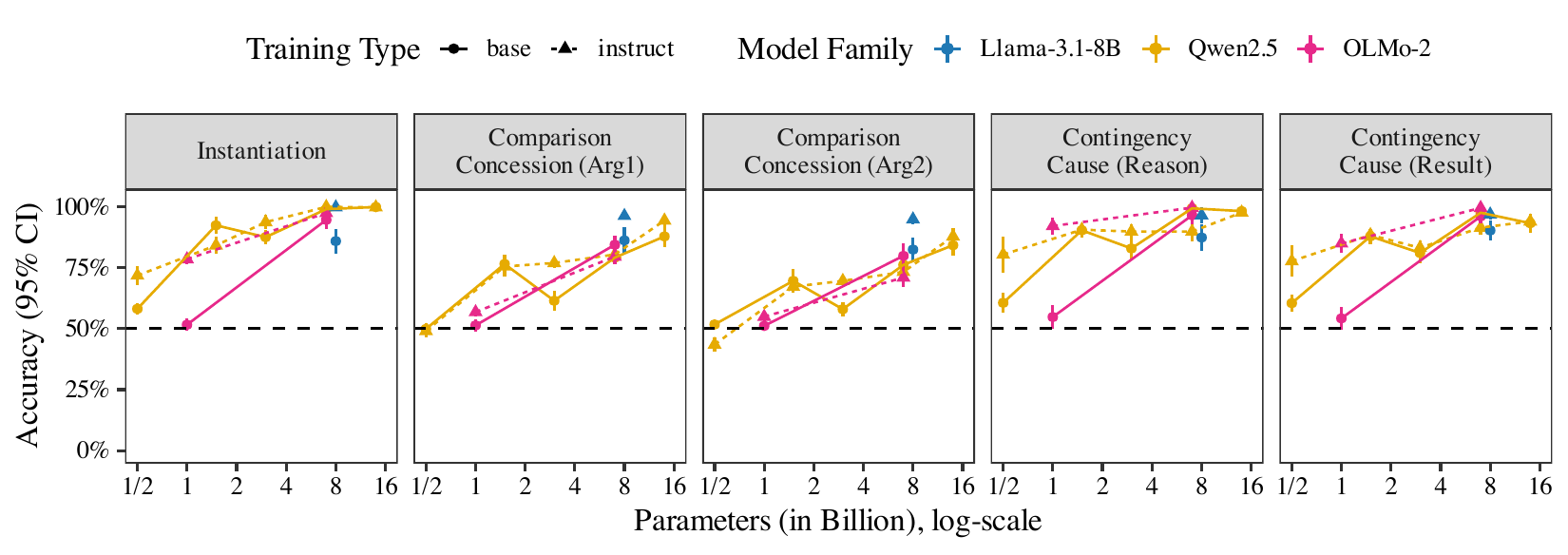}
    \vspace{-1em}
    \caption{Accuracy of LMs across connective senses, where stimuli matches world knowledge. 
    The black dashed line indicates chance performance (50\%). Error bars indicate 95\% confidence intervals measured across connectives and prompt variation.
    }
    \label{fig:world_knowledge}
    \vspace{-0.5em}
\end{figure*}

\paragraph{Instruction Tuning} Instruction-tuned models barely showed any improvements over their base counterparts. There were scattered exceptions to this trend---e.g., OLMo 2 1B on \texttt{Contingency} connectives, Qwen-2.5-500M on \texttt{Precedence} connectives (though the opposite effect was found in \texttt{Succession}), most Qwen-2.5 models and OLMo 2 7B on \texttt{Succession} connectives. However, due to the absence of a consistent pattern, it is not yet clear if instruction tuning shows any particular benefit in the context of making inferences from discourse connective usages. An LMER analysis using an interaction between sense and instruction tuning (coded as a binary fixed-effect---1 if present, 0 otherwise), to predict accuracy, with random effects for prompt template and model yields results consistent with this conclusion ($\beta_{\text{instruct}} = -0.004, p = .826$).

\paragraph{Reasoning-based Tuning} While we did not have as many cases of ``Reasoning-based'' tuning (in the sense of DeepSeek-R1) as we did Instruction Tuning, we do find evidence---preliminary though it might be---of reasoning-based tuning improving models' abilities to make novel entity inferences from discourse connectives. The \lrm{} model consistently outperformed its base and instruction tuning variants in all except the concession class of connectives.  A linear-mixed effects regression analysis on results from all three variants of Qwen 2.5 14B (base, instruct, reasoning), with an interaction between training type and sense, and random effects of prompt template yielded corroborating evidence ($\beta_{\text{reasoning vs. base}} = 0.06, p < .001; \beta_{\text{reasoning vs. instruct}} = 0.08, p < .001$).

\section{Post-hoc Experiment: Effect of World Knowledge}

Is the poor performance of the model on preference (concessive and comparison) and instantiation stimuli due to an inability to reason about these connectives in general, or only when it is necessary to use that reasoning to infer information about nonce words? To test this, we run an additional experiment with a similar set of stimuli, where nonce words are replaced with real entities. As a result, our questions simply ask about inferences already grounded in world knowledge, without requiring explicit reasoning. Therefore, any weaknesses here could further shed light on the fundamental limitations of models on connectives.

\paragraph{Stimuli Design} The stimuli for this second experiment matches the forms described in \cref{sec:stimuli_design}, and uses the same properties laid out in \Cref{tab:stats}. The key difference is the use of real-world entities instead of nonce words, matching the licensed entailments with world knowledge. As examples, see \cref{ex:grounded_pref} and \cref{ex:grounded_inst} for preference and instantiation stimuli respectively. See \Cref{sec:appendix-world-knowledge} for a full list of all entities and associated properties.
\ex. \label{ex:grounded_pref}
 \label{ex:g_pref} \textit{Because} I love fruits, I prefer mangos to kale.

\ex. \label{ex:grounded_inst}
\label{ex:g_inst} I like birds. \textit{For example}, sparrows are nice.

\paragraph{Results}

With world knowledge, model performance generally increased with model size across all connective senses, as seen in \Cref{fig:world_knowledge}. 
This contrasts sharply with experiments using nonce words, where we did not find effects of scale. 
Despite overall higher performance, there is still room for improvement on concessive connectives in particular, and especially for models under 8B parameters. This indicates general difficulties on this particular family of connectives, even when world knowledge reinforces the inferences they license. 
That said, large models' significant increases in performance in all categories demonstrates the value of our approach using nonce words, as it is able to highlight the ways models fail to grasp entailments licensed by connectives in novel contexts specifically.

\section{Conclusion}

We present \dataset{}, a dataset that sheds light on LMs' ability to reason about their knowledge of discourse connectives in order to make inferences about novel entities. In doing so, we complement a long-standing body of work that has primarily treated connectives---rather than world knowledge---as the main target of prediction. While we pursued a number of different analyses, our most salient conclusion was about \textit{all} tested models' systematic failure on \textit{concession} connectives. In addition, we failed to find any effect of connective frequency, scaling, or even instruction tuning on LMs' performance, though there was preliminary evidence in favor of reasoning-based tuning, opening up future analyses of the linguistic properties of ``reasoning models'' \citep{guo2025deepseek}.
Overall, we advocate for more investigations into the functional meanings of even the most simplest of linguistic cues, to fully catalog how language exposure shapes the learning of meaning.

\section{Limitations}

There are a number of limitations to this work:

\paragraph{Sole reliance on connective} While our aim in this work is to isolate---to the extent that we can---the reliance of LMs to the connectives alone and make inferences about the ``world'', our stimuli are not fully devoid of non-connective meaning. That is, LMs must still have to critically rely on the meanings of other words in the context (e.g., `love', 'hate', 'prefer' in the case of preference stimuli). Fully teasing apart non-connective meanings is nearly impossible, so we relied on the assumption that models can consistently reason over these.

\paragraph{Gradability of properties}
Another key assumption we've made in the preference stimuli is that the properties in question are non-gradable (to argue that we've captured entailments rather than implicatures---which are much easily cancelable). While this might be true for certain properties (e.g., \textit{is an island, is a college town}, it is up for debate if any property is truly binary. At the same time, there is active discussion about whether category membership and ``goodness of exemplar''/typicality (the metric people use to argue that a property is graded, e.g., a \textit{platypus} is less ``mammal-like'' than \textit{bear}) are separate attributes of categories \citep{hampton2007typicality}, with category membership being more binary-like than typicality.

\paragraph{How novel are our novel words?}
Since we are using surface forms whose subtokens exist in the LMs' vocabulary, it is difficult to truly deem the nonce word as a ``novel'' word. One solution is to insert novel tokens, though using them requires one to fine-tune the model, which makes our analysis especially intractable, since we do not necessarily have contexts to train the model on. Furthermore, we have counterbalanced our nonce words in our stimuli such that any particular bias a model might have picked up on will result in chance level performance. 
We have additionally verified that model do not necessarily have a systematic bias towards a single nonce word in our stimuli, ruling out the possibility that the nonce words themselves are the reason for the poor performance on concessive connectives.

\paragraph{Caveats in model comparison}

 A caveat to the  takeaway that reasoning models like \lrm{} are better is that it is non-trivial to directly compare across the three classes (base, instruct, reasoning), since the \lrm{} is expected to produce a set of reasoning traces before generating its output whereas models in the other two classes are directly queried for answers.

 \paragraph{Larger models} While we do not test larger models, we have been careful in drawing conclusions in our work, sticking to the set of models tested here. Since \dataset{} is model agnostic, we will open our benchmark to anyone who wishes to test a model on it. We consider 17 models, with the kinds of variance in properties we have, to be a reasonable number of models to test.

\section*{Acknowledgments}
The inception of this project was inspired by Yufang Hou and her keynote ``Bridging Resolution: A Journey Towards Modeling Referential Discourse Entities'' at CODI 2023, which contained real-world examples of LMs' failures on discourse connectives.
We thank Kyle Mahowald and Sebastian Schuster for their input at the initial stages of this work. We are especially grateful to Yanai Elazar for making the Dolma frequencies available to us. We thank the three anonymous reviewers, and especially reviewer FCRj for their engagement with our submission. We are also grateful to Bonnie Webber for her detailed comments on an earlier version. This work was partially supported by NSF grants IIS-2107524, IIS-2145479 and a grant from Open Philanthropy. The authors acknowledge the Texas Advanced Computing Center (TACC)\footnote{\url{https://tacc.utexas.edu}} at The University of Texas at Austin for providing computational resources that have contributed to the research results reported within this paper. Kanishka Misra was supported by the Donald D. Harrington Faculty Fellowship at UT Austin.

\bibliography{custom}

\clearpage
\appendix

\section{Detailed Prompt Templates}
\label{sec:prompt-templates}

\Cref{tab:prompt-variation-pref-inst} shows detailed prompt templates as well as other prompt artifacts (premise and inference examples, names) for the preference and instantiation stimuli. \Cref{tab:prompt-variation-temporal} shows that information for temporal stimuli.

\begin{table*}[ht]
\centering
\resizebox{\textwidth}{!}{%
\begin{tabular}{@{}ll@{}}
\toprule
\textbf{\texttt{Name} Options} & Emily, Lucy, Adam, John, Cameron, Erica, Megan, David, Jessica, Daniel\\ \midrule
\textbf{\begin{tabular}[c]{@{}l@{}}\texttt{premise}\\ Example\end{tabular}} &
  \begin{tabular}[c]{@{}l@{}}\textbf{Preference:} Although I hate islands, I prefer wugsville to daxburgh.\\ \textbf{Instantiation:} I like wugs, for example, daxes are nice.\end{tabular} \\ \midrule
\textbf{\begin{tabular}[c]{@{}l@{}}\texttt{inference}\\ Example\end{tabular}} &
  \begin{tabular}[c]{@{}l@{}}\textbf{Preference:} wugsville is an island\\ \textbf{Instantiation:} daxes are wugs\end{tabular} \\ \midrule
\multirow{12}{*}{\textbf{Templates}} &
  \texttt{\{Name\}} said, ``\texttt{\{premise\}}'' From this, is it true that \texttt{\{inference\}}? \texttt{\{Y/N instruction\}} \\
 & \texttt{\{Name\}} said, ``\texttt{\{premise\}}'' Does this mean that \texttt{\{inference\}}? \texttt{\{Y/N instruction\}}                               \\
 & \texttt{\{Name\}} said, ``\texttt{\{premise\}}'' Can we conclude from this that \texttt{\{inference\}}? \texttt{\{Y/N instruction\}}                    \\
 & \texttt{\{Name\}} said, ``\texttt{\{premise\}}'' Does this suggest that \texttt{\{inference\}}? \texttt{\{Y/N instruction\}}                            \\
 & \texttt{\{Name\}} said, ``\texttt{\{premise\}}'' Can we say from this that \texttt{\{inference\}}? \texttt{\{Y/N instruction\}}                         \\
 & \texttt{\{Name\}} said, ``\texttt{\{premise\}}'' Can we conclude from what \texttt{\{Name\}} said that \texttt{\{inference\}}? \texttt{\{Y/N instruction\}} \\
 & \texttt{\{Name\}} said, ``\texttt{\{premise\}}'' Can we say from what \texttt{\{Name\}} said that \texttt{\{inference\}}? \texttt{\{Y/N instruction\}}      \\
 & \texttt{\{Name\}} said, ``\texttt{\{premise\}}'' Does \texttt{\{Name\}} mean that \texttt{\{inference\}}? \texttt{\{Y/N instruction\}}                      \\
 & \texttt{\{Name\}} said, ``\texttt{\{premise\}}'' Does what \texttt{\{Name\}} said suggest that \texttt{\{inference\}}? \texttt{\{Y/N instruction\}}         \\
 & \texttt{\{Name\}} said, ``\texttt{\{premise\}}'' If you heard this, would you think that \texttt{\{inference\}}? \texttt{\{Y/N instruction\}}           \\
 & \texttt{\{Name\}} said, ``\texttt{\{premise\}}'' If you heard \texttt{\{Name\}}, would you think that \texttt{\{inference\}}? \texttt{\{Y/N instruction\}}  \\
 & \texttt{\{Name\}} said, ``\texttt{\{premise\}}'' From what \texttt{\{Name\}} said, do you think that \texttt{\{inference\}}? \texttt{\{Y/N instruction\}}   \\ \bottomrule
\end{tabular}%
}
\caption{Prompt variation for \textbf{Preference} and \textbf{Instantiation} Stimuli. In all cases, the variable \texttt{\{Y/N instruction\}} is always ``Answer either with Yes or No.''}
\label{tab:prompt-variation-pref-inst}
\end{table*}

\begin{table*}[]
\centering
\resizebox{\textwidth}{!}{%
\begin{tabular}{@{}ll@{}}
\toprule
\textbf{\texttt{Name} Options} &
  Emily, Lucy, Adam, John, Cameron, Erica, Megan, David, Jessica, Daniel \\ \midrule
\textbf{\texttt{event} Options} &
  Wugfest, Daxday, Gextravaganza, Blicketbash, Fepfestival \\ \midrule
\textbf{\begin{tabular}[c]{@{}l@{}}\texttt{premise}\\ Example\end{tabular}} &
  \texttt{event1} occurred. Thereafter, \texttt{event2} took place. \\ \midrule
\multirow{12}{*}{\textbf{Templates}} &
  \texttt{\{Name\}} said, ``\texttt{\{premise\}}" From this, which event started first? \texttt{\{Answer Instruction\}} \\
 &
  \texttt{\{Name\}} said, ``\texttt{\{premise\}}" From this, which event started earlier? \texttt{\{Answer Instruction\}} \\
 &
  \texttt{\{Name\}} said, ``\texttt{\{premise\}}" From this, which event began first? \texttt{\{Answer Instruction\}} \\
 &
  \texttt{\{Name\}} said, ``\texttt{\{premise\}}" From this, which event began earlier? \texttt{\{Answer Instruction\}} \\
 &
  \texttt{\{Name\}} said, ``\texttt{\{premise\}}" From this, which of the two events began first? \texttt{\{Answer Instruction\}} \\
 &
  \texttt{\{Name\}} said, ``\texttt{\{premise\}}" From this, which of the two events began earlier? \texttt{\{Answer Instruction\}} \\
 &
  \texttt{\{Name\}} said, ``\texttt{\{premise\}}" From what \texttt{\{Name\}} said, which event started first? \texttt{\{Answer Instruction\}} \\
 &
  \texttt{\{Name\}} said, ``\texttt{\{premise\}}" From what \texttt{\{Name\}} said, which event started earlier? \texttt{\{Answer Instruction\}} \\
 &
  \texttt{\{Name\}} said, ``\texttt{\{premise\}}" From what \texttt{\{Name\}} said, which event began first? \texttt{\{Answer Instruction\}} \\
 &
  \texttt{\{Name\}} said, ``\texttt{\{premise\}}" From what \texttt{\{Name\}} said, which event began earlier? \texttt{\{Answer Instruction\}} \\
 &
  \texttt{\{Name\}} said, ``\texttt{\{premise\}}" From what \texttt{\{Name\}} said, which of the two events began first? \texttt{\{Answer Instruction\}} \\
 &
  \texttt{\{Name\}} said, ``\texttt{\{premise\}}" From what \texttt{\{Name\}} said, which of the two events began earlier? \texttt{\{Answer Instruction\}} \\ \bottomrule
\end{tabular}%
}
\caption{Prompt variation for \textbf{Temporal} stimuli. \texttt{\{Answer Instruction\}} is always ``Answer either with \texttt{event1} or \texttt{event2} and nothing else.''}
\label{tab:prompt-variation-temporal}
\end{table*}

\section{Model Metadata}
\label{sec:modelmeta}

\Cref{tab:model-metadata} shows details about each model in this work.

\begin{table*}[]
\centering
\small
\resizebox{\textwidth}{!}{%
\begin{tabular}{@{}llrl@{}}
\toprule
\textbf{Family} & \textbf{\texttt{HuggingFace Identifier}} & \textbf{Parameters (in billion)} & \textbf{Training Type} \\ \midrule
\multirow{2}{*}{Llama-3.1-8B} & \texttt{meta-llama/Meta-Llama-3.1-8B} & 8 & base \\
 & \texttt{meta-llama/Meta-Llama-3.1-8B-Instruct} & 8 & instruct \\ \midrule
\multirow{10}{*}{Qwen2.5} & \texttt{Qwen/Qwen2.5-0.5B} & 0.5 & base \\
 & \texttt{Qwen/Qwen2.5-0.5B-Instruct} & 0.5 & instruct \\
 & \texttt{Qwen/Qwen2.5-1.5B} & 1.5 & base \\
 & \texttt{Qwen/Qwen2.5-1.5B-Instruct} & 1.5 & instruct \\
 & \texttt{Qwen/Qwen2.5-3B} & 3 & base \\
 & \texttt{Qwen/Qwen2.5-3B-Instruct} & 3 & instruct \\
 & \texttt{Qwen/Qwen2.5-7B} & 7 & base \\
 & \texttt{Qwen/Qwen2.5-7B-Instruct} & 7 & instruct \\
 & \texttt{Qwen/Qwen2.5-14B} & 14 & base \\
 & \texttt{Qwen/Qwen2.5-14B-Instruct} & 14 & instruct \\ \midrule
Qwen2.5-DS & \texttt{deepseek-ai/DeepSeek-R1-Distill-Qwen-14B} & 14 & reasoning \\ \midrule
\multirow{4}{*}{OLMo-2} & \texttt{allenai/OLMo-2-0425-1B} & 1 & base \\
 & \texttt{allenai/OLMo-2-0425-1B-Instruct} & 1 & instruct \\
 & \texttt{allenai/OLMo-2-1124-7B} & 7 & base \\
 & \texttt{allenai/OLMo-2-1124-7B-Instruct} & 7 & Instruct \\ \bottomrule
\end{tabular}%
}
\caption{Family, Huggingface identifier, parameter counts, and training type for LMs evaluated in this work.}
\label{tab:model-metadata}
\end{table*}

\section{Implementation details}
\label{sec:implementation}
All models' log-probabilities were extracted using \texttt{minicons} \citep{misra2022minicons}, on a cluster with 4 NVIDIA A40 GPUs. Most experiments were run on a single A40 GPU, with the exception of the 14B Qwen models and their reasoning versions (run on two).

\section{Breakdown by Individual Connective}
\label{sec:breakdown}

\begin{figure*}
    \centering
    \includegraphics[width=\linewidth]{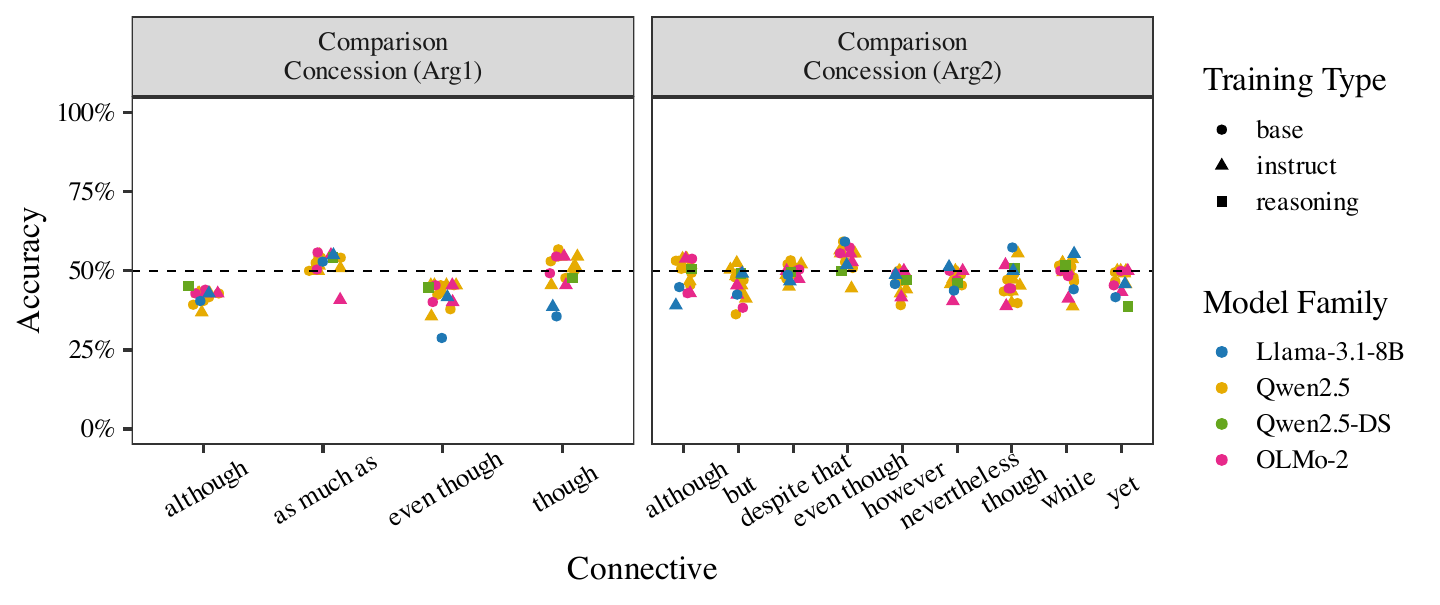}
    \caption{Results on Comparison connectives.}
    \label{fig:comparison-breakdown}
\end{figure*}

\begin{figure*}
    \centering
    \includegraphics[width=\linewidth]{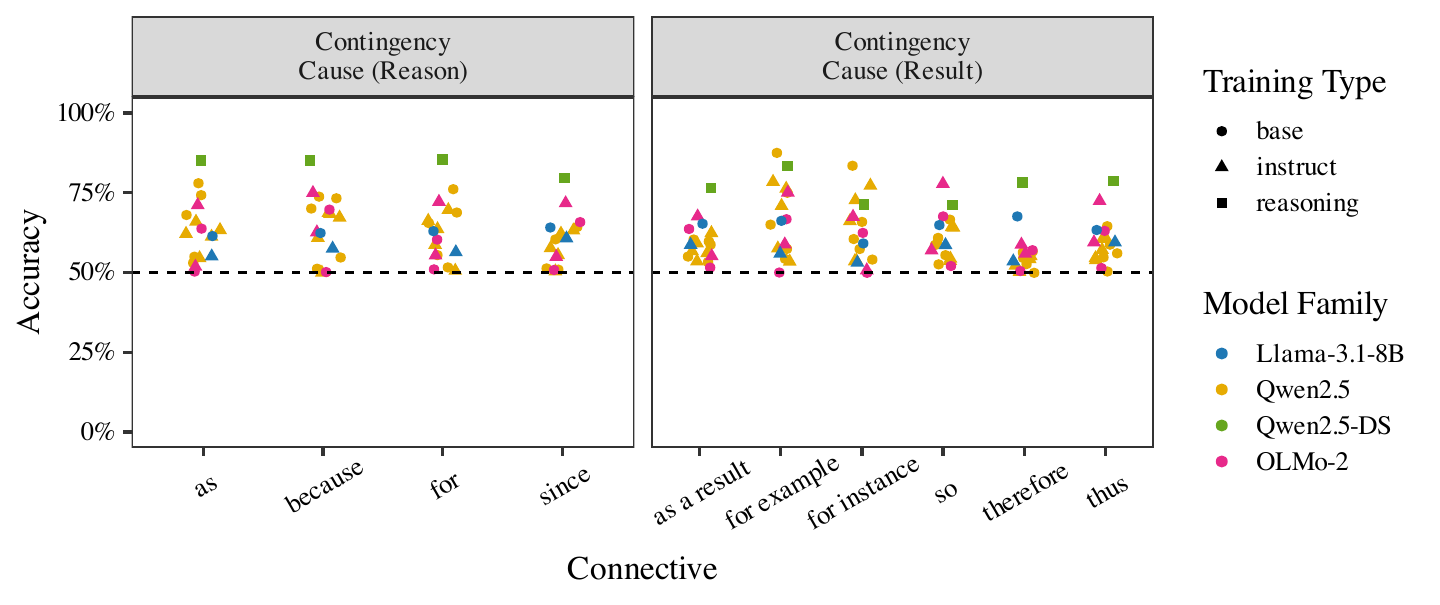}
    \caption{Results on Contingency connectives.}
    \label{fig:contingency-breakdown}
\end{figure*}

\begin{figure*}
    \centering
    \includegraphics[width=\linewidth]{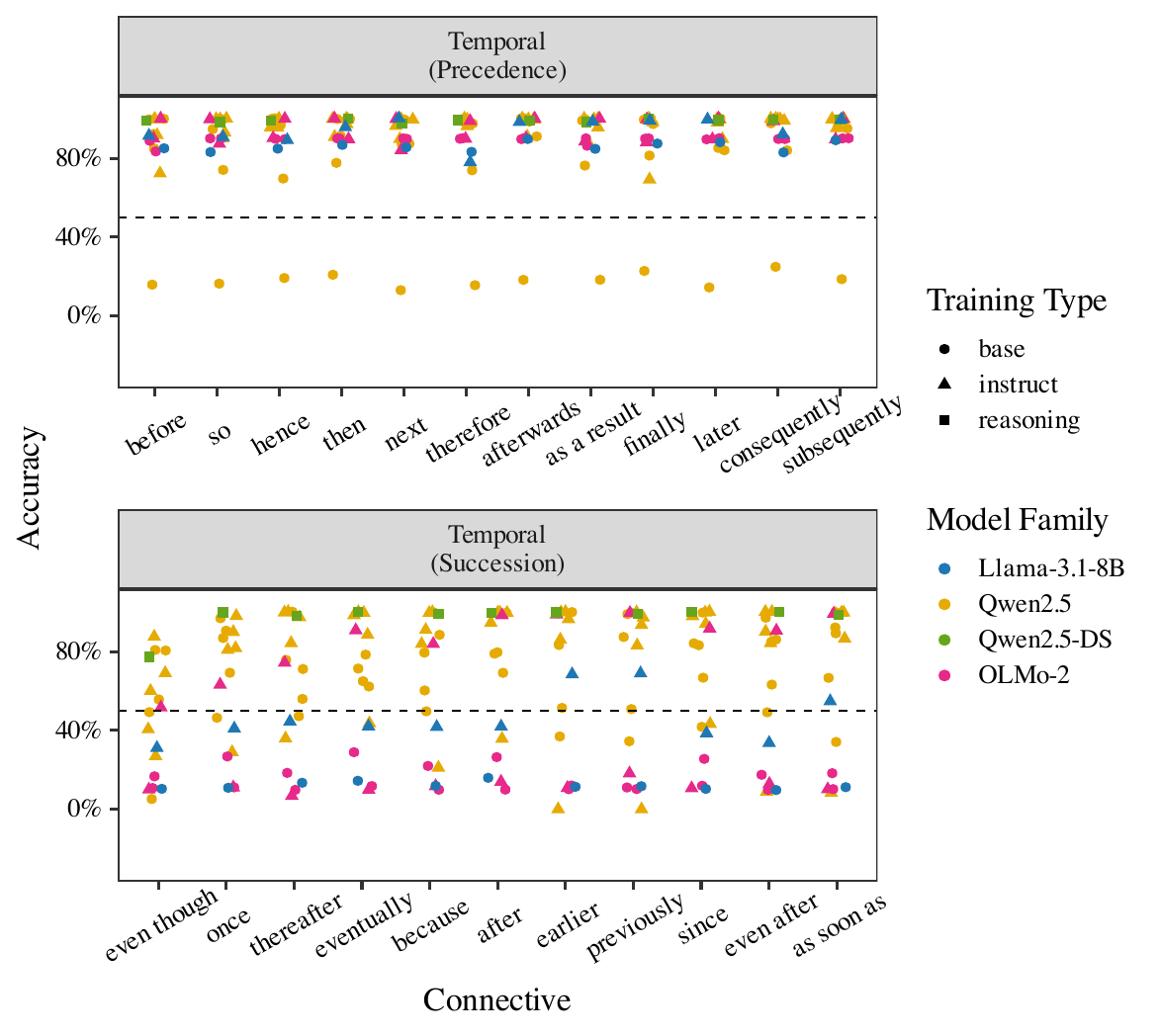}
    \caption{Results on Temporal connectives.}
    \label{fig:temporal-breakdown}
\end{figure*}

We show plots for model results broken down per connective in this section. \Cref{fig:inst} shows results for \sense{Expansion.Instantiation} (main text); \Cref{fig:comparison-breakdown} shows results for \sense{Contingency.Cause}; \Cref{fig:contingency-breakdown} shows results for \sense{Comparison.Concession}; and \Cref{fig:temporal-breakdown} shows results for \sense{Temporal} connectives.

\section{Frequency Analysis}
\label{sec:frequency}

\Cref{tab:freq-r-squared} shows Pearson's correlation between models' accuracies and connective frequencies. We find no evidence of positive correlation for any model.

\begin{table}[]
\centering
\small

\begin{tabular}{@{}lrr@{}}
\toprule
\textbf{model} & \textbf{$r^2$} & \textbf{$p$} \\ \midrule
Qwen2.5-0.5B & -0.08 & 0.61 \\
Qwen2.5-0.5B-Instruct & 0.10 & 0.53 \\
Qwen2.5-1.5B & -0.02 & 0.91 \\
Qwen2.5-1.5B-Instruct & -0.19 & 0.25 \\
Qwen2.5-14B & -0.16 & 0.33 \\
Qwen2.5-14B-Instruct & -0.21 & 0.19 \\
Qwen2.5-3B & -0.14 & 0.38 \\
Qwen2.5-3B-Instruct & -0.26 & 0.11 \\
Qwen2.5-7B & -0.05 & 0.75 \\
Qwen2.5-7B-Instruct & -0.27 & 0.10 \\
OLMo-2-1B & 0.09 & 0.57 \\
OLMo-2-1B-Instruct & 0.14 & 0.40 \\
OLMo-2-7B & 0.15 & 0.37 \\
OLMo-2-7B-Instruct & -0.22 & 0.18 \\
Qwen2.5-DS & -0.15 & 0.37 \\
Llama-3.1-8B & 0.15 & 0.36 \\
Llama-3.1-8B-Instruct & -0.08 & 0.62 \\ \bottomrule
\end{tabular}%
\caption{Correlation ($r^2$) between LMs' connective-level accuracy and connective frequency, as estimated from the Dolma corpus \citep{soldaini-etal-2024-dolma}. $p$ denotes p-values.}
\label{tab:freq-r-squared}
\end{table}

\section{Linear Mixed-Effects Regression analysis}
\label{sec:lmer}
We describe our linear mixed-effects regression analysis to understand the effect of Scale, Instruction Tuning, and Reasoning-based Tuning on results. All analyses were conducted using the \texttt{lmerTest} library to specify the model, and the \texttt{car} library to perform significance tests for interaction effects. 

\paragraph{Scale} We divide the parameter counts (\texttt{params}) by \texttt{1e+9} and use the following formula, where \texttt{sense} specifies the sense of the connectives: 
\begin{align*}
    \texttt{accuracy} &\sim \texttt{params} \times \texttt{sense} + (1 \mid \texttt{family})\\
    &+ (1 \mid \texttt{prompt\_template}) 
\end{align*}

\paragraph{Instruction tuning} We code the instruct variable to be 1 if the model was instruction-tuned and 0 otherwise, discarding the reasoning model (\lrm{}) from this analysis (since it was only applicable for one model class). We use the following formula:
\begin{align*}
    \texttt{accuracy} &\sim \texttt{instruct} \times \texttt{sense} + (1 \mid \texttt{model})\\
    &+ (1 \mid \texttt{prompt\_template}) 
\end{align*}

\paragraph{Reasoning-based Tuning} We perform this analysis only for the Qwen2.5-14B base class, and sum-code the \texttt{training_mode} variable (base, instruct, reasoning). We use the following formula:
\begin{align*}
    \texttt{accuracy} \sim& \texttt{training\_mode} \times \texttt{sense}\\ &+ (1 \mid \texttt{prompt\_template})
\end{align*}

\section{Algorithmic extraction of reasoning model responses}
\label{sec:appendix-response-extraction}

In all cases, we set the model temperature to be 0.6, as recommended in its model card (\url{https://huggingface.co/deepseek-ai/DeepSeek-R1-Distill-Qwen-14B}). After extracting the model's generations, we perform the following preprocessing steps.

\begin{enumerate}
    \item For consistency, remove all of the following characters: * \textbackslash{}n ' (  ). 
    \item Remove all instances of the phrases ``answer:'' and ``the answer is''
    \item If \texttt{boxed\textbackslash{}\{ } or \texttt{boxed\{ } is in the text, return the answer within curly braces.
    \item Set the trace to all lowercase.
    \item Determine if the trace is a temporal answer or not by checking for the presence of any of the temporal nonce words, or two misspellings (``blicktash'', ``bicketbash'')
    \item For non-temporal traces: If the words ``yes'' or ``no'' appear in the first or last three or two characters respectively, return that as the answer.
    \item For temporal traces: return the nonce word that appears either as the first or last $n$  characters, where $n$ is the length of the nonce word in characters.
\end{enumerate}

\section{Real-world entities} 
\label{sec:appendix-world-knowledge}
The real properties and entities used in the second experiment on the effects of world knowledge can be found in \Cref{tab:grounded_properties} and \Cref{tab:grounded_properties_inst} for preference and instantiation respectively.

\begin{table*}[!th]
\centering
\small
\begin{tabular}{@{}lll@{}}
\toprule
\multicolumn{3}{c}{\textbf{Preference}} \\ \midrule
\textbf{Property} & \textbf{Entities with property} & \textbf{Entities without property} \\ \midrule
Leafy vegetables & \textit{\begin{tabular}[c]{@{}l@{}}kale, spinach, lettuce, \\ arugula, cabbage\end{tabular}} & \textit{\begin{tabular}[c]{@{}l@{}}asparagus, cucumbers,\\  tomatos, bell peppers, onions\end{tabular}} \\ \cmidrule(l){2-3} 
Mammals & \textit{\begin{tabular}[c]{@{}l@{}}mammals, dogs, cats,\\  sheep, cows, pigs\end{tabular}} & \textit{\begin{tabular}[c]{@{}l@{}}lizards, frogs, snakes,\\  turtles, toads\end{tabular}} \\ \cmidrule(l){2-3} 
Fruits & \textit{\begin{tabular}[c]{@{}l@{}}fruits, grapes, apples, \\ mangos, pears, strawberries,\end{tabular}} & \textit{\begin{tabular}[c]{@{}l@{}}kale, spinach, lettuce, \\ arugula, cabbage\end{tabular}} \\ \cmidrule(l){2-3} 
String instruments & \textit{\begin{tabular}[c]{@{}l@{}}violins, violas, cellos, \\ double basses, guitars\end{tabular}} & \textit{\begin{tabular}[c]{@{}l@{}}trumpets, trombones,\\  clarinets, flutes, saxaphones\end{tabular}} \\ \cmidrule(l){2-3} 
Insects & \textit{\begin{tabular}[c]{@{}l@{}}flies, beetles, grasshoppers,\\  cicadas, ants\end{tabular}} & \textit{mice, bat, gerbils, dogs, cats} \\ \cmidrule(l){2-3} 
College towns & \textit{\begin{tabular}[c]{@{}l@{}}Lubbock, Texas; \\ Ann Arbor, Michigan;\\  Oxford, England; \\ Cambridge, England\end{tabular}} & \textit{\begin{tabular}[c]{@{}l@{}}London, England;\\ Dallas, Texas;\\ Detroit, Michigan;\\ Liverpool, England\end{tabular}} \\ \cmidrule(l){2-3} 
Has mountains nearby & \textit{\begin{tabular}[c]{@{}l@{}}Salt Lake City, Utah; \\ Telluride, Colorado; \\ Tucson, Arizona; \\ Jackson Hole, Wyoming;\\  Palm Springs, California\end{tabular}} & \textit{\begin{tabular}[c]{@{}l@{}}Dallas, Texas; \\ Chicago, Illinois; \\ Jacksonville, Florida; \\ New Orleans, Louisiana;\\ Houston, Texas\end{tabular}} \\ \cmidrule(l){2-3} 
Has an equatorial climate & \textit{Singapore} & \textit{\begin{tabular}[c]{@{}l@{}}Toronto, Canada; \\ Oslo, Norway; \\ Moscow, Russia\end{tabular}} \\ \cmidrule(l){2-3} 
Is coastal & \textit{\begin{tabular}[c]{@{}l@{}}Miami, Cancun, Boston,\\  Lisbon, Mumbai,\end{tabular}} & \textit{\begin{tabular}[c]{@{}l@{}}Dallas, Mexico City, Reno, \\ Pittsburgh, Madrid, Delhi\end{tabular}} \\ \cmidrule(l){2-3} 
Is an island & \textit{\begin{tabular}[c]{@{}l@{}}Galveston, Puerto Rico,\\ Sicily, Japan, Hawaii\end{tabular}} & \textit{\begin{tabular}[c]{@{}l@{}}Oklahoma, Nevada, Brazil, \\ South Korea, Spain\end{tabular}} \\ \bottomrule
\end{tabular}
\caption{Properties used for preference stimuli in the second experiment grounded in world knowledge.}
\label{tab:grounded_properties}
\end{table*}

\begin{table}[]
\centering
\small
\begin{tabular}{@{}ll@{}}
\toprule
\multicolumn{2}{c}{\textbf{Instantiation}} \\ \midrule
\textbf{Set} & \textbf{Instance} \\ \midrule
\textit{Birds} & \textit{Sparrows, ravens, crows} \\
\textit{Mammals} & \textit{Cats, dogs, sheep} \\
\textit{Fruits} & \textit{Grapes, apples, pears} \\
\textit{String instruments} & \textit{Violins, cellos, violas,} \\
\textit{Insects} & \textit{Beetles, grasshoppers, ants} \\
\textit{Fish} & \textit{Tuna, salmon, trout} \\ \bottomrule
\end{tabular}
\caption{Properties used for instantiation stimuli in the second experiment grounded in world knowledge.}
\label{tab:grounded_properties_inst}
\end{table}

\section{License}
We release \dataset{} with an MIT license.

\end{document}